\documentclass{bmvc2k}

\title{OctreeNCA: Single-Pass 184 MP Segmentation on Consumer Hardware}

\addauthor{Nick Lemke}{nick.lemke@tu-darmstadt.de}{1}
\addauthor{John Kalkhof \\ Niklas Babendererde \\ Anirban Mukhopadhyay}{}{1}

\addinstitution{
 Technical University of Darmstadt,\\
 Karolinenplatz 5,\\
 64289 Darmstadt, Germany
}

\def\etal{\emph{et al}\bmvaOneDot}

\runninghead{Lemke \etal}{OctreeNCA}

\usepackage{array,multirow,graphicx}
\usepackage{float}
\usepackage{todonotes}
\usepackage{xcolor}
\usepackage{colortbl}
\usepackage[]{algorithm2e}

\usepackage{adjustbox}
\usepackage{siunitx}

\usepackage{listings}
\definecolor{backcolour}{rgb}{0.95,0.95,0.92}
\definecolor{codegreen}{rgb}{0,0.6,0}

\usepackage{mathrsfs}
\newcommand{\loss}{\mathscr{L}}

\definecolor{tabbestcolor}{rgb}{0.004, 0.141, 0.337}

\def \best {\cellcolor{tabbestcolor!30}}
\def \sbest {\cellcolor{tabbestcolor!15}}

\usepackage[capitalise]{cleveref}
\usepackage{booktabs}

\begin{document}

\maketitle

\begin{abstract}
Medical applications demand segmentation of large inputs, like prostate MRIs, pathology slices, or videos of surgery. These inputs should ideally be inferred at once to provide the model with proper spatial or temporal context. 
When segmenting large inputs, the VRAM consumption of the GPU becomes the bottleneck. 
Architectures like UNets or Vision Transformers scale very poorly in VRAM consumption, resulting in patch- or frame-wise approaches that compromise global consistency and inference speed.
The lightweight Neural Cellular Automaton (NCA) is a bio-inspired model that is by construction size-invariant. However, due to its local-only communication rules, it lacks global knowledge. We propose OctreeNCA by generalizing the neighborhood definition using an octree data structure. Our generalized neighborhood definition enables the efficient traversal of global knowledge. Since deep learning frameworks are mainly developed for large multi-layer networks, their implementation does not fully leverage the advantages of NCAs. We implement an NCA inference function in CUDA that further reduces VRAM demands and increases inference speed.  
%Taking inspiration from computer graphics, we leverage an octree structure to handle global knowledge efficiently. 
%Combined with the lightweight bio-inspired Neural Cellular Automata we propose OctreeNCA capable of segmenting high-resolution images and long videos in one go.
%Combined with our custom CUDA NCA inference function, 
Our OctreeNCA segments high-resolution images and videos quickly while occupying 90\% less VRAM than a UNet during evaluation. This allows us to segment 184 Megapixel pathology slices or 1-minute surgical videos at once.
Our implementation is publicly available: \url{https://github.com/MECLabTUDA/OctreeNCA}.
\end{abstract}

\section{Introduction}

% Medical images and videos should be inferred at once to give spatial + temporal context and allow the model to predict smooth transitions between frames of a video or 

Medical image analysis often deals with large, high-resolution images, enabling high-precision diagnosis and treatments. Ideally, these images should be processed at once, so the model can leverage long-range context of 2D and 3D images. For example, when segmenting surgical videos, this would mean merging the frames into a 2D+t volume.

Starting with the well-known UNet~\cite{ronneberger2015u}, recent state-of-the-art deep-learning models have become progressively bigger, putting more and more demands on the GPU. Not only does segmentation with larger models take longer, but it also requires more memory on the GPU (VRAM). 
For example, the auto-ML framework nnUNet~\cite{isensee2019nnu} specifies a minimum of 4 GiB of VRAM for inference. The larger the model, the worse they scale towards inference on larger inputs with manageable VRAM demands (\cref{fig:raspi_speed_and_gpu_vram}). 
%For example, the nnUNet framework~\cite{isensee2019nnu} specifies a minimum of 4 GB of VRAM for training a UNet segmentation model. The larger the model, the worse they scale towards training on larger inputs. Also, inference has to be performed on a similar resolution as the model was trained on, to maintain high-quality predictions.

This massive increase in VRAM demands increases the divide in access to healthcare worldwide. On one hand, there are high-income countries, that have many trained physicians and enough financial resources to acquire hardware capable of segmenting large images encountered in the clinical routine. On the other hand, there are low- and middle-income countries (LMICs) that lack well-trained physicians, that could benefit most from AI assistance~\cite{guo2018application}. However, LMICs also lack the infrastructure for deploying a model on high-resolution images~\cite{bellemo2019artificial}.

The empirical memory footprint with respect to the input size can be seen in \cref{fig:raspi_speed_and_gpu_vram}. The UNet~\cite{ronneberger2015u} is a high-quality segmentation model.
Although the VRAM consumption of a UNet scales linearly, the scaling factor is too large to conveniently scale to large images. Also, UNets can fail when evaluated on larger images than those seen during training~\cite{kalkhof2023med}. Transformer-based models~\cite{xie2021segformer, dosovitskiy2020image} can be applied to an arbitrary image size with equally good segmentation quality. The problem with vision transformers is that the VRAM demands scale quadratically with the number of input pixels/voxels. 
The TransUNet~\cite{chen2021transunet} is a combination of a vision transformer and an UNet, capable of high-quality segmentation. It outperforms the standard UNet and Transformer-based baselines in terms of segmentation accuracy. The downside of the TransUNet is the high number of parameters and the even higher VRAM demand during inference. Foundation models are even larger models trained on gigantic amounts of data. The segment anything model (SAM)~\citep{kirillov2023segment} is a foundation model with more than 500 Million parameters capable of segmenting natural or medical images~\cite{ranem2023exploring}. Despite its backbone consisting of a vision transformer, SAM internally resizes the image to $1024\times1024$ pixel, which limits its segmentation quality on large images.

A completely different architecture is the bio-inspired Neural Cellular Automaton (NCA) \cite{mordvintsev2020growing}. The NCA is a lightweight architecture that learns the rules of a cellular automaton capable of generating complex patterns with simple rules. The NCA has been used to segment natural images~\cite{sandler2020image}, as well as, medical images~\cite{kalkhof2023med,kalkhof2023m3d}. By limiting the communication of NCAs to a 1-pixel neighborhood, the NCA has a much lower memory footprint during evaluation than UNets or Transformers.
However, the downside of this local-only communication is that many iterations of NCA rules are needed to propagate global context. This makes inference and training unnecessarily slow.

Inspired by the \emph{vision as inverse graphics} idea, we utilize the quadtree/octree representation for orchestrating and managing a large scene. Specifically, we propose \textbf{OctreeNCA which generalizes the definition of a neighborhood} by handling long-range dependencies on a high level of the octree while still maintaining the regular grid of pixels/voxels. Since traditional NCAs are implemented in deep learning frameworks developed for layer-wise networks~\cite{paszke2019pytorch}, NCA inference is not as efficient as it could be. Hence, we implement a \textbf{custom CUDA layer for quick and efficient NCA inference} on the GPU.
Our OctreeNCA performs segmentation much quicker than other deep networks while reducing VRAM demands during training and evaluation. We show OctreeNCA's efficacy on three different medical segmentation tasks. In the prostate MRI segmentation setting, we are dealing with $320\times320\times24$, which is more than $2.5$ Million voxels. Even larger than that, are pathology slices. Scanned at $10\times$ magnification (pixel resolution \SI{0.48}{\micro\meter}) they have a size of $23500\times52500$, which is equal to a $1.22$ Gigapixel image. Videos of cholecystectomy surgeries have a spatial resolution of $240\times432$ captured at a frame rate of $25$ frames per second. The surgery videos in our dataset are at least $30$ minutes long, resulting in $4.7\times 10^9$ spatio-temporal voxels for a single video.

Our OctreeNCA segments 184 Megapixel pathology images and 1-minute videos with large-scale spatial/temporal context in a few seconds on a single inference step. Such an inference with other SOTA methods like Transformers or UNets is not even remotely possible due to much higher VRAM demands.
Our OctreeNCA opens the possibility of mitigating the societal differences between clinics in LMICs that only have access to cheap/outdated hardware and developed countries with access to large-scale hardware.

% Prosate: 320x320x24 > 2.5 Megavoxel
% Pathology: 23742 x 52644 > 1.24 Gigapixel
% Cholec: 240 x 432 > 100k pixel (this is convenient low resolution, but no temporal context!) with temporal context of 3 seconds (80 frames) -> 8.3 Million Pixel

\section{Related work}
% general introduction to segmentation models for medical imaging
% different methods to segmenting these larger inputs
% different applications of NCAs and specially image segmentation 
For medical image segmentation, different deep-learning architectures are used, which we review here. Depending on the anatomy of interest / the segmentation task, different approaches are leveraged to handle the large-scale inputs in medical imaging. 
Also, we highlight some of the related works using Neural Cellular Automata, especially in the context of (medical) image segmentation.

\subsection{Medical Image Segmentation}
%go into detail on patch by patch segmentation of pathology slices

% UNet & nnUNet -> patchwise inference due to risk of edge cases that could lead to bad performance

% khened2021generalized: non-overlapping patch by patch (256x256) on a low resolution for pN-staging (no segmentation). use a three model ensemble
% xu2017large: patchwise feature extraction + classification SVM to create segmentation at the resolution of the patches
% feng2021deep: multiresolution segmentation + majority voting

% -> those methods are slow! especially when overlapping patches for transferring information between patches
\begin{figure*}[t]
    \centering
    \includegraphics[width=\linewidth]{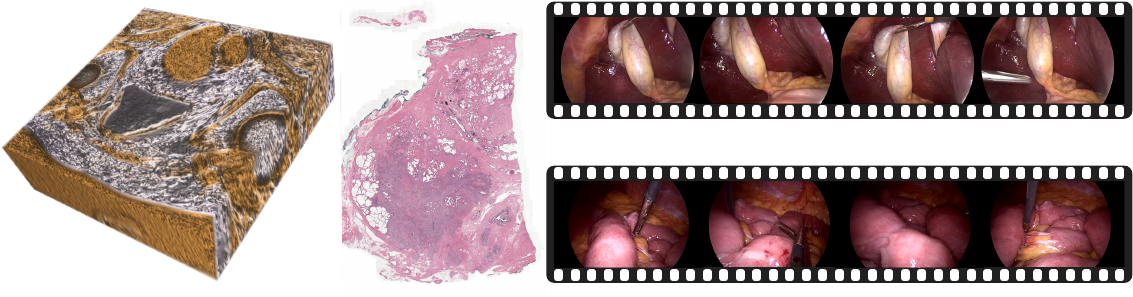}
    \caption{Volume-rendering of a high-resolution prostate MRI (left), a pathology image (middle), and a cholec surgery video (right).}
    \label{fig:task_examples}
\end{figure*}

Medical image segmentation is dominated by fully convolutional networks like the UNet~\cite{ronneberger2015u}. Enveloped by a preprocessing pipeline and a network and training planning module, the nnUNet framework~\cite{isensee2019nnu} sets the gold standard in medical image segmentation of 2D and 3D radiology images. Inspired by natural language processing, the vision transformer (ViT)~\cite{dosovitskiy2020image} splits an image into patches and encodes a sequence of patches into expressive features. The architecture of transformers contains self-attention layers that allow the traversal of information from any patch to any other patch. Hence, the ViT performs well at capturing long-range dependencies.
The TransUNet~\cite{chen2021transunet} combines a transformer encoder and a UNet decoder to build a strong segmentation model. A problem with transformer-based architecture is that their runtime and VRAM demands scale quadratically with the input resolution (\cref{fig:raspi_speed_and_gpu_vram}).

Large inputs arise in histopathology, radiology, and in surgical videos. Histopathology images have a resolution in the Gigapixel regime. MRI and CT images are 3D images leading to a high number of voxels. A video stacked to a 2D+t tensor may result in an even larger input depending on the video length and framerate.
Examples of the inputs of the three segmentation tasks can be seen in \cref{fig:task_examples}.

\subsection{Neural Cellular Automata}
A novel architecture that by design is size and translation-invariant is the neural cellular automaton (NCA). 
An initial work of generating an image from a single seed~\cite{mordvintsev2020growing} shows that NCAs can learn complex patterns with very few parameters (compared to other recent deep architectures).

%NCAs are applied to generative modeling tasks using a variational setup~\cite{palm2022variational}, a generative adversarial setup~\cite{otte2021generative}, or in a denoising diffusion modeling scenario~\cite{kalkhof2024frequency,elbatel2024organism}.

Also, NCAs are used for image segmentation on low-resolution $96\times96$ images~\citep{sandler2020image} due to high VRAM demands during training. 
As medical image segmentation requires models to be applied to higher resolutions, the two-level Med-NCA is proposed. The first NCA diffuses global knowledge on $1/4$ of the original resolution. After that, the state channels are upscaled to the original resolution, where a second NCA performs the final segmentation~\citep{kalkhof2023med}. This two-level approach allows for training on $320\times320$ images with a VRAM consumption of only $6.8$ GiB during training. %The authors show that the segmentation performance does not depend on the resolution of the inputs~\cite{kalkhof2023med}.
Some of these methods are evaluated on images that have a different size from those they are trained on, showing promise of size-invariance of NCAs~\cite{kalkhof2024frequency, kalkhof2023med}.

Our OctreeNCA utilizes a relaxed definition of the 1-pixel neighborhood at different octree levels. While other NCAs need many steps to diffuse global knowledge, our OctreeNCA handles global knowledge with ease by leveraging the first octree level to distribute context across the entire image in a few steps.

\section{Methodology}

% Divide in
%   NCA (NCA are scale-independent segmentation networks) 
%   Octree NCA (Easy training and inference using OctreeNCA)
%   Octree NCA in nnUNet (OctreeNCA as a plug-and-play module in the nnUNet framework)

In this section, we present the NCA backbone architecture used in our work. After that, we highlight our OctreeNCA and clarify why training an OctreeNCA is much easier than a traditional NCA. In the last part, we shed light on our custom CUDA implementation and elucidate why OctreeNCA works so much faster than other architectures.

\begin{figure*}
    \centering
    \includegraphics[width=\linewidth]{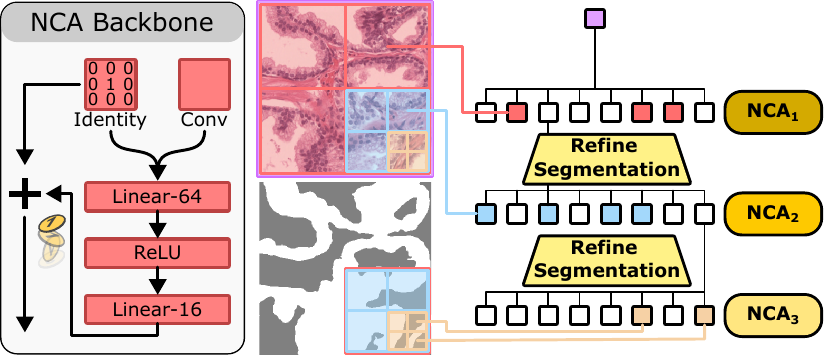}
    \caption{Left: Our backbone NCA architecture. Right: Our OctreeNCA architecture. For visualization purposes with three levels only.}
    \label{fig:octree_architecture}
\end{figure*}

\subsection{NCAs for Size-invariant Segmentation}
Inspired by cellular automata such as Conway's Game of Life a new architecture known as neural cellular automata (NCA) is proposed~\cite{mordvintsev2020growing}. Like classical cellular automata, NCAs are repeatedly applied to the same input, changing the state of cells using its update rule. However, unlike classical cellular automata, the update rules are not hand-engineered but learned by a neural network~\citep{mordvintsev2020growing}.

Each cell of the NCA consists of a multi-dimensional vector. The first $n$ channels are occupied by the input image. For example, when segmenting RGB images, the first $3$ channels are filled with the input, yielding an initial seed for the NCA iterations. The other channels, known as hidden channels, are left to the NCA to store and distribute knowledge as learned by the neural network. The NCA is repeatedly applied to each cell on the grid. The NCA does not have any information on the surrounding cells outside of its $3\time3$ neighborhood, which makes the NCA invariant to the input size by design.

The NCA consists of a convolution, two linear layers, and a ReLU activation. 
Each channel of the $3\times3$ neighborhood is convolved with the learned kernel. The output of the convolution is concatenated with the current state of the cell and fed through the linear, ReLU, linear network, computing an additive update. We adopt the asynchronous activation mechanism from previous works~\cite{kalkhof2023m3d,kalkhof2023med,mordvintsev2020growing,sandler2020image} by randomly setting the additive update to $0$ by a $50\%$ chance. 

Different from previous architectures~\cite{kalkhof2023m3d}, we did not employ batch normalization in the NCA because we noticed that the segmentation performance was dependent on online-batch statistics during evaluation~\cite{sandler2020image}. Also, restraining from a normalization layer reduces memory demands during training. Furthermore, a global normalization layer like batch norm ruins the local-only property of (neural) cellular automata, which would disable the implementation of a custom CUDA kernel we present in \cref{sec:cuda_layer}.
%Since we did not find improvements neither in terms of training convergence nor in segmentation quality, we decided to not use any normalization layer, yielding the quickest training convergence and \todo{best} segmentation quality.
The architecture of our backbone NCA can be found in~\cref{fig:octree_architecture}.

\subsection{Efficient Training of OctreeNCA}
Training NCAs on large-scale inputs puts a high demand on the VRAM capacity of the GPU. Hence, we propose OctreeNCA, which reduces the high VRAM consumption during training, allowing us to train NCAs on clinically-relevant precision. The OctreeNCA architecture is illustrated in \cref{fig:octree_architecture}.

\textbf{Repeated hidden channel upscaling:}
Our OctreeNCA takes a full-resolution input image and builds an octree/quadtree structure by averaging pixels in each node. After that, a backbone NCA operates on the highest level of the octree (lowest resolution images) to diffuse global knowledge quickly. In the next step, the global knowledge encoded in the hidden channels of the NCA grid is upsampled using nearest-neighbor interpolation and transferred to the next level of the octree. A second NCA is applied to the second level of the octree. Since this NCA can access global knowledge through the hidden channels from the previous level \textbf{as few as 10 steps are enough to refine} the segmentation to the higher resolution of the octree. This hidden channel upscaling and segmentation refining procedure is repeated until a final NCA predicts the segmentation mask in its last hidden channel on the lowest octree level (original image resolution).
We train the entire OctreeNCA end-to-end to allow individual NCAs to encode any knowledge capable of solving the segmentation task without imposing any constraints. The NCAs share the same architecture, but each has its own weights.

\textbf{Reduced VRAM demands during Training:}
Training the OctreeNCA has a much lower VRAM memory footprint, which is thanks to two advantages we will elaborate on in the following:
1) Fewer NCA steps are needed to diffuse global knowledge and achieve high-resolution segmentation. Due to the local-only communication of NCAs, the model needs at least $\max(H,W)$ steps to transfer knowledge from one side of an $H\times W$ large image to the other side. Thanks to our OctreeNCA, we can reduce the number of steps by transferring global knowledge at $1/2^L$ of the original resolution, needing only $\max(H,W)/2^L$ steps to achieve the same, where $L$ is the number of levels in the octree.
2) Most of our NCAs are applied to lower input resolutions. During training, we need to store the intermediate results of the NCA for each cell and each step to do the backpropagation efficiently. Hence, the VRAM demands during training scale with the input size of our NCA. In our OctreeNCA, only the last NCA operates on the full image resolution, which in turn means that the VRAM requirements on the higher-level NCAs are much more moderate.

\subsection{Custom NCA-tailored CUDA Kernel for Large-scale Segmentation}\label{sec:cuda_layer}
Since NCAs are by design size-invariant, we can segment much larger images than the model has seen during training. Although this is technically possible with other models such as UNets~\cite{ronneberger2015u} or Transformers~\cite{dosovitskiy2020image,xie2021segformer}, their VRAM demands during inference scale worse than those of NCA-based methods (see \cref{fig:raspi_speed_and_gpu_vram}).

Also, the typical implementation of NCAs in deep-learning frameworks~\cite{paszke2019pytorch, abadi2016tensorflow} developed for layer-wise architectures like CNNs or Transformers does not fully leverage the benefits of the lightweight NCA. The layer-wise implementation computes one layer of the backbone NCA across the entire image at the same time, which results in large intermediate results after computing the NCA hidden representation (Linear-64 in \cref{fig:octree_architecture}). A more convenient implementation for Cellular Automata takes the input image and computes an entire NCA step without splitting it into different layers. The hidden representation of the NCA is small enough to fit into the local memory of recent CUDA cores, so a cell-oriented (instead of layer-oriented) implementation does not require more VRAM than the initial input seed and the corresponding output.

We implement a CUDA kernel that uses pre-trained NCA weights to perform the NCA inference using a cellular automata-inspired implementation. Our implementation further reduces the VRAM demands of NCA inference. Also, inference is carried out a lot quicker compared to other methods like UNets and NCA-based architectures. The inference acceleration is not only due to our specialized CUDA implementation but also because 1) fewer NCA steps are needed to provide accurate segmentations with global context and 2) on average, the backbone NCAs in our OctreeNCA operate on lower resolution than those in the other NCA baselines~\cite{kalkhof2023m3d, kalkhof2023med}.

\begin{figure}
    \centering
    \includegraphics[width=\linewidth]{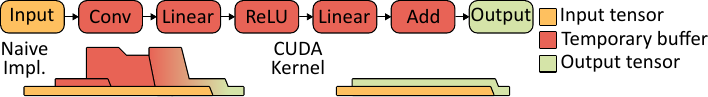}
    \caption{Naive layer-wise NCA implementation (left). Our NCA CUDA implementation (right). The histograms show the VRAM allocations during inference. Our kernel eliminates the need to allocate temporary buffers.}
    \label{fig:cuda_implementation}
\end{figure}

Since VRAM demands scale linearly with the number of cells (pixels), we investigate the VRAM demand for a single cell. We visualize our analysis in \cref{fig:cuda_implementation}. The traditional layer-wise implementation has a maximum VRAM demand of 96 channels $\times$ the number of pixels/cells. A thorough explanation of the Figure can be found in the supplementary material.

In comparison, our CUDA implementation merges all operations into a single CUDA kernel, storing the 96 intermediate results in the thread-local memory. So only 16 input and 16 output channels have to be allocated at once.
In \cref{sec:supp_cuda_details} in our supplementary material, we provide further details, including a pseudo-code implementation of our inference function.

% prostate segmentation minimal VRAM
% prostate segmentation nnUNet framework
% video segmentation (measure run time!)
% whole-slide/6 Megapixel pathology inference
% pathology extrapolation

% Datasets

% Implementation

\section{Datasets}
For each dataset, we create a test split containing $30\%$ of the patients in the dataset. The other $70\%$ are used to train the models.

% Prostate: medical segmentation decathlon
% resize to 320x320x24 (median resolution: ???)
\textbf{Radiology:} We use prostate data originating from the medical segmentation decathlon~\cite{antonelli2022medical}. It consists of 32 T2-weighted MRIs. The segmentation mask encompasses both the peripheral and transition zone. The median resolution is $320\times320\times20$. We follow previous works~\cite{kalkhof2023m3d} and resize the MRIs to $320\times320\times24$. 

% Epithelium segmentation: PESO dataset
% 
\textbf{Pathology:} For the pathology segmentation task we use the publicly available PESO dataset~\cite{bulten2019epithelium}. We use a subset featuring $30$ whole-slide images of H\&E stained radical prostatectomy slides. The labels encompassing the epithelial cells have been generated with AI assistance. The slices are scanned with a pixel resolution of \SI{0.24}{\micro\meter}. However, we downsample the resolution to pixel resolution \SI{0.48}{\micro\meter} to make the dataset more manageable in size. The downsampled slides have a resolution of $23742\times52644$. We train the models on $320\times 320$ patches.% For the sake of simplicity, we evaluate our models on the same patch size. 

% Cholec: CholecSeg8k based on unlabeled cholec80

\textbf{Surgery:} We use the CholecSeg8k dataset~\cite{hong2020cholecseg8k} which is a densely segmented subset of the Cholec80 dataset~\cite{twinanda2016endonet}. The dataset features $80$ frame-long videos of dense annotations. In total, there are $101$ video snippets of $17$ different cholecystectomy surgeries. The videos are captured at a resolution of $854\times480$. We resize the videos to $432\times240$ for the ease of training and testing. The dataset features $12$ classes of which seven are underrepresented in the dataset. To simplify the framework we refrain from balancing techniques and remove the underrepresented classes from the segmentation masks. We leave the abdominal wall, liver, fat, grasper, and gallbladder classes in our dataset. To test our model on longer video sequences we use the Cholec80 dataset~\cite{twinanda2016endonet}, which does not provide segmentation masks.

\section{Results}
In this section, we empirically investigate the capabilities of our OctreeNCA. First, we compare OctreeNCA to other methods quantitatively and qualitatively. After that, we show the inference speed on a \$50 Raspberry Pi. Further results are found in the appendix, which contains results on large-scale images and videos~(\cref{sec:results:large_scale}), an ablation study~(\cref{sec:results:ablation}), detailed quantitative~(\cref{sec:results:detailed_quantitative}) and further qualitative results~(\cref{sec:results:more_qualitative}).

\subsection{Quantitative Comparisons}

\begin{figure}[t]
\centering
\includegraphics[width=\linewidth]{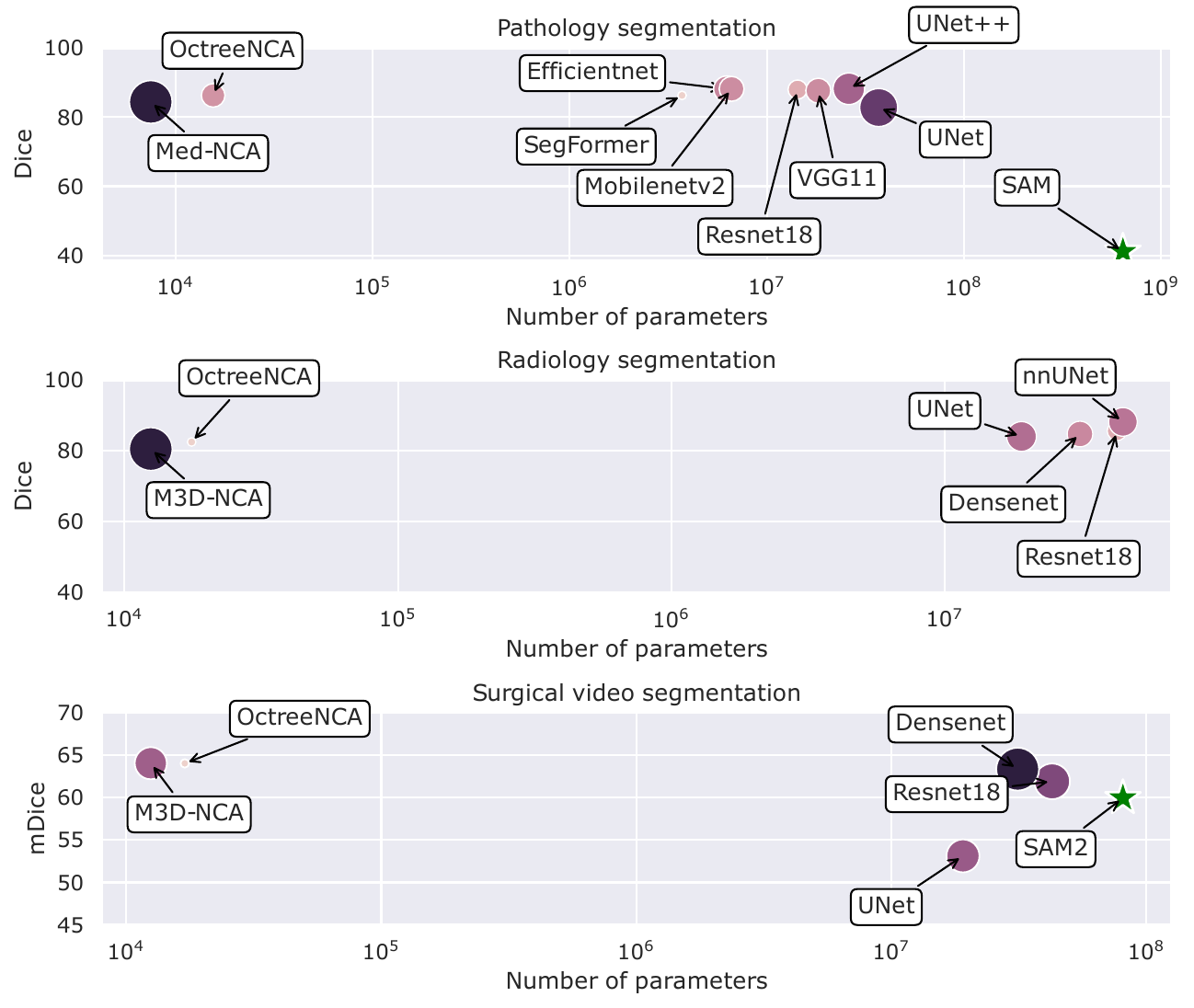}
\caption{Dice score on epithelium and prostate segmentation, and mean dice (mDice) on cholec segmentation. Circle size represents VRAM demands during training.}
\label{fig:all_three_seg}
\end{figure}

\Cref{fig:all_three_seg} shows qualitative comparisons of our OctreeNCA with other segmentation architectures tailored for the respective domain. Across the three segmentation tasks, our OctreeNCA performs similarly well to UNet baselines while having much fewer parameters. Also, the VRAM demands are much smaller than those of the other models. The exact values can be found in the supplementary material.

\subsection{Qualitative Comparisons}
\begin{figure*}[htb]
    \centering
    \includegraphics[width=\linewidth]{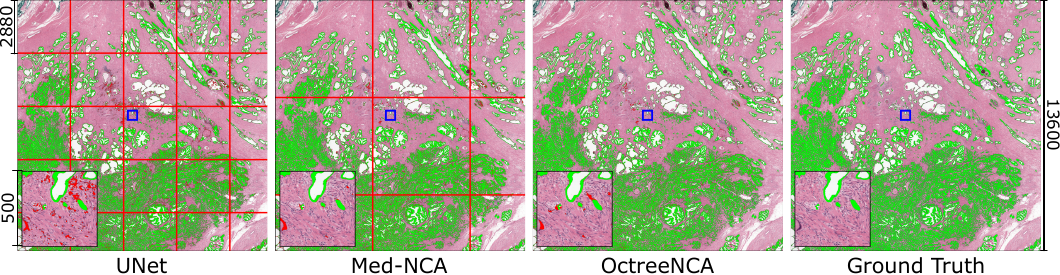}
    \caption{Pathology segmentation. True positive pixels are green, false positives are red. The red grid indicates the patch-wise inference of the UNet and Med-NCA. The inset shows a zoomed-in version of the blue patch.}
    \label{fig:peso_qualitative}
\end{figure*}

\Cref{fig:peso_qualitative} shows qualitative results on a 148 MP pathology image. True positive pixels are indicated in green, and false positives are red. The UNet has to be applied in patches, indicated by the red grid. Our OctreeNCA segments the entire image at once while providing better segmentation quality than the UNet. We provide additional qualitative results in the supplementary material.

\subsection{Inference Speed on Raspberry Pi}
\begin{figure}[htb]
    \centering
    \includegraphics[width=0.49\linewidth]{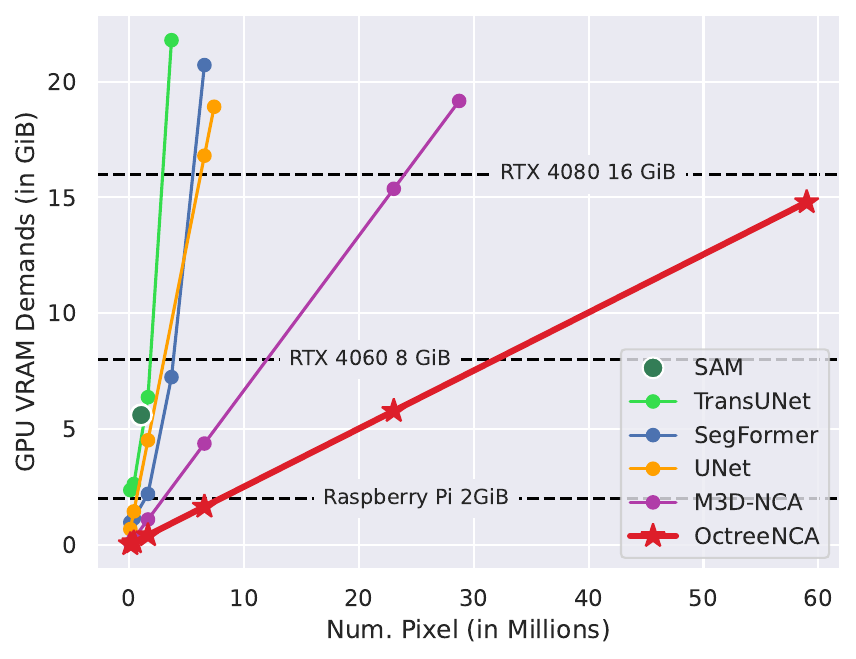}
    \includegraphics[width=0.49\linewidth]{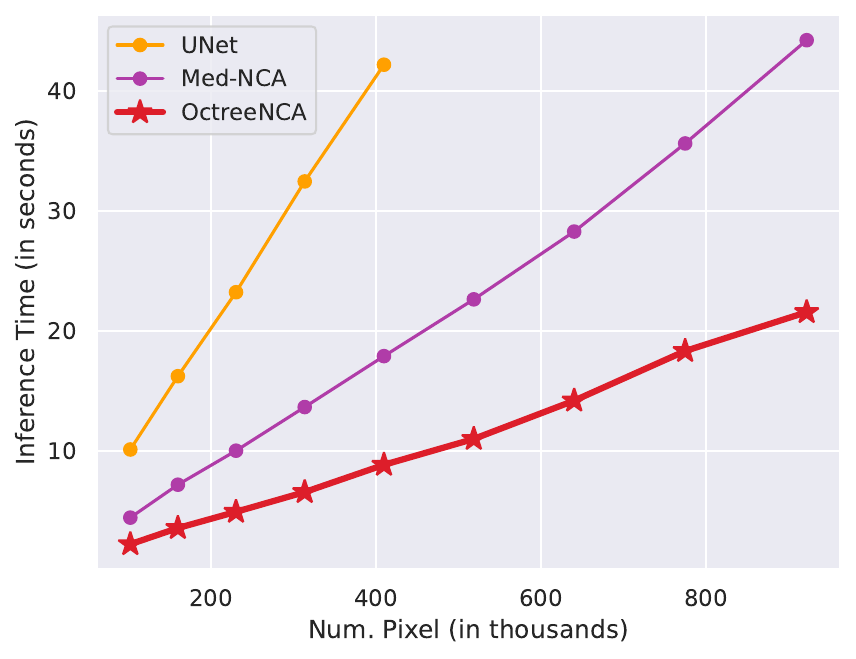}
    \caption{Left: The empirical memory footprint on the VRAM of different architectures during inference on different-sized images. OctreeNCA scales the best.
    Right: Inference runtime on a Raspberry Pi 4 Model B 2GiB. The runtime of OctreeNCA scales much better than those of UNet and Med-NCA.}
    \label{fig:raspi_speed_and_gpu_vram}
\end{figure}

\Cref{fig:raspi_speed_and_gpu_vram} shows the inference time with respect to the number of pixels in the input on a \$50 Raspberry Pi 4 Model B with 2GiB RAM. The inference time using our OctreeNCA scales by a lower factor than that of the UNet and Med-NCA. Also, the UNet runs out of VRAM when segmenting images larger than 0.41 Megapixel. The exact values are found in the supplementary material.

\section{Conclusion}
In this work, we propose OctreeNCA a new architecture based on NCA that simplifies the training and evaluation of high-resolution data by managing long-range dependencies in an octree data structure. We present a novel CUDA layer to make use of the minimalistic architecture of the NCA.
We show OctreeNCA's capabilities in terms of VRAM demands, number of parameters, and inference speed on three different segmentation tasks in the fields of pathology, radiology, and surgery. 

The VRAM demands during inference are significantly lower than those of other models thanks to our custom CUDA layer and the octree structure managing the global context. Even on very cheap devices like a \$50 Raspberry Pi, OctreeNCA demonstrates a clear advantage to larger models by reducing the inference time while segmenting larger images. 

We look forward to seeing OctreeNCA adapted for a large variety of tasks that go beyond segmentation or even beyond image analysis. Especially in the healthcare sector, we hope to close the gap between developing countries that cannot afford excessive hardware for training and inference, and developed countries capable of training gigantic foundation models on hundreds of GPUs.

\section*{Acknowledgments}
This work has been partially funded by the Federal Ministry of Education and Research project ”FED-PATH” (grant 01KD2210B) and the Software Campus project ”FedVS4Hist” (grant 01|S23067).

\bibliography{egbib}

\newpage

\clearpage

\section{Literature Review on Medical Image Segmentation}
Whole-slide pathology images have a resolution in the Gigapixel regime of $10^{10}$ pixels or even more~\cite{lotz2016patch}, which is infeasible to process at once with a neural network at the current state of the hardware. Hence, deep models are evaluated by segmenting one patch at a time and stitching the results together~\cite{khened2021generalized,xu2017large,feng2021deep}. A patch-by-patch segmentation is not ideal, due to the increase in input-output-operations, which slows down the segmentation of large images. Also, the edges of the patches lack global knowledge resulting in a reduction in segmentation accuracy.

Even though the radiology images are much smaller compared to whole-slide pathology images, they are still high-dimensional data, because of the third dimension added by MRIs or CTs. The nnUNet framework~\cite{isensee2019nnu} trains a backbone UNet on 3D patches to reduce the memory footprint during training. Evaluation using the nnUNet framework is performed patch-by-patch with the same patch size that was used during training to reduce the threat of edge effects when changing the patch size for inference.

In video segmentation, we deal with a series of frames recorded during surgery. To allow the segmentation model to access temporal knowledge we consider the set of frames as a 3-dimensional volume~\cite{hara2018can,tran2015learning}. Similar to segmenting a 3D dimensional radiology image, 3D convolutions or 3D NCAs can build a strong baseline for video segmentation with spatiotemporal context.

\section{Space Complexity of Segmentation Algorithms}%{Theoretical Analysis of VRAM demands}

As fully convolutional networks are theoretically speaking size and translation-invariant they can be applied to high-resolution images, the only limit being the memory availability. In this section, we provide an analysis of the space complexity of UNets, Transformers, and NCAs.

\textbf{UNet}: Due to the multi-resolution skip connections of a UNet its VRAM requirements scale according to $\mathcal{O}(c_0\cdot d + c_1\cdot d/2 + \dots)$ where $d$ is the number of pixels/voxels in our input and $c_i$ is the number of channels stored on the $i$-th resolution. A common design decision is to choose $c_{i+1}=2\cdot c_i$ \cite{ronneberger2015u, isensee2019nnu} which simplifies the expression to $\mathcal{O}( N \cdot c_0\cdot d)$ where $N$ is the number of resolutions in the UNet. The originally proposed UNet uses $N=5$ and $c_0=64$ \cite{ronneberger2015u} which results in a linear scaling with a factor of $320$.

\textbf{Transformer:} Transformer-based architectures rely on the self-attention mechanism. Self-attention computes a pairwise attention score, resulting in a quadratic run time and space complexity $\mathcal{O}(d^2)$~\cite{keles2023computational}. To reduce the amortized run time and space demands, the image is split into tokens. For example, the vision transformer~\cite{dosovitskiy2020image} splits the input image into $16\times16$ sized patches. This procedure reduces the practical runtime and space demands by introducing a small factor to the complexity: 
\begin{align}
    \mathcal{O}\left(\left(\frac{d}{4}\right)^2\right) = \mathcal{O}\left(\frac{1}{16}d^2\right)\notag
\end{align}
This trick does not change the space complexity leaving transformers with a quadratic space complexity. 

\textbf{NCA:}
In theory, the NCA has a space complexity of $\mathcal{O}(c\cdot d)$ where $c$ is the dimensionality of the cells (we choose $c=16$) and $d$ is the number of pixels/cells. As we have shown the naive practical implementation has a much higher VRAM demand of over $\mathcal{O}(144\cdot d)$. However, we manage to reduce it to only $\mathcal{O}(32\cdot d)$. We provide an analysis of how to get these numbers in \cref{sec:vram_analysis} in the supplementary.

\section{Implementation Details of our NCA CUDA Inference Function}\label{sec:supp_cuda_details}
% https://en.wikipedia.org/wiki/CUDA#Version_features_and_specifications
% https://developer.download.nvidia.com/CUDA/training/register_spilling.pdf

% number of input channels and hidden size of the NCA must be known at compile time -> compiler can optimize the code like unrolling loops, and allocating defining storage for the thread-local arrays.
% no big deal, compilation is easy and a few different variants could be compiled at once to features a variety of NCA backbones
% alternatively, Just-in-time compilation could be used for variable parameters

Our CUDA and C++ code for the improved NCA inference is publicly available in our source code: \url{https://github.com/MECLabTUDA/OctreeNCA}. A pseudo-code implementation can be found in Listing~\ref{lst:cuda_pseudo}. In this section, we provide a few insights into the implementation and design decisions of our kernel.

The number of channels and the hidden size of the NCA must be known at compile time. We take this limitation so the compiler can optimize the code by unrolling loops and defining the storage for the thread-local arrays. However, hard-coding the channels and the hidden size of the NCA is no big deal, as there are multiple simple solutions to this problem if necessary: One can compile different versions of the code featuring any combination of the number of channels and hidden sizes.
Alternatively, even more flexibility can be achieved by compiling the function just-in-time after specifying the number of channels and the hidden size of the NCA. All NCAs had the same architecture in our experiments, so we used a single pre-compiled inference function.

% Our implementation does not feature the segmentation of batches. If necessary, the function can be extended rather easy, for example, by iterating over an additional dimension

Our implementation does not allocate the output on its own but takes a pre-allocated tensor in its arguments. This has the advantage, that we can give pre-sampled random variables to the kernel. For the random fire functionality, we need random variables. To keep the pseudo-random forward pass comparable to the Pytorch implementation,  we decided to leave Pytorch's random number generator and feed the random variables in the output tensor — that way our implementation delivers the same results as the traditional Pytorch implementation. The implementation of how to invoke the NCA inference function can be found in \cref{lst:inference_wo_func}.

Computing the results in place, and writing the outputs to the input tensor would not work, because that would require synchronization between the CUDA cores, which comes with a large overhead.

% even for larger models the output tensor can be used to store up to C temporary values in the VRAM
\begin{lstlisting}[caption=CUDA kernel of our inference function., label={lst:cuda_pseudo}, float=tp]
# CUDA kernel is invoked for each pixel with coordinates (x, y) on the GPU simultaneously
def nca_forward_kernel_2d(state: float[], x: int, y: int, result: float[]):
	# define local buffer of fixed size
	local_arr: float[64]
	# convolution centered around (x, y)
	local_arr = Conv(state, (x,y))
	# proceed with inference, put intermediate results in local buffer, do not write to VRAM
	local_arr = Concat(state[x,y], local_arr)
	local_arr = ReLU(Lin(local_arr))
	# compute next NCA state and store in result
	result[x,y] = Bernoulli() * Lin(local_arr) + state[x,y]
\end{lstlisting}

\section{Detailed VRAM Analyses of Layer-wise NCA Implementation vs. our CUDA Implementation}
\label{sec:vram_analysis}
\begin{figure*}
    \centering
    \includegraphics[width=\linewidth]{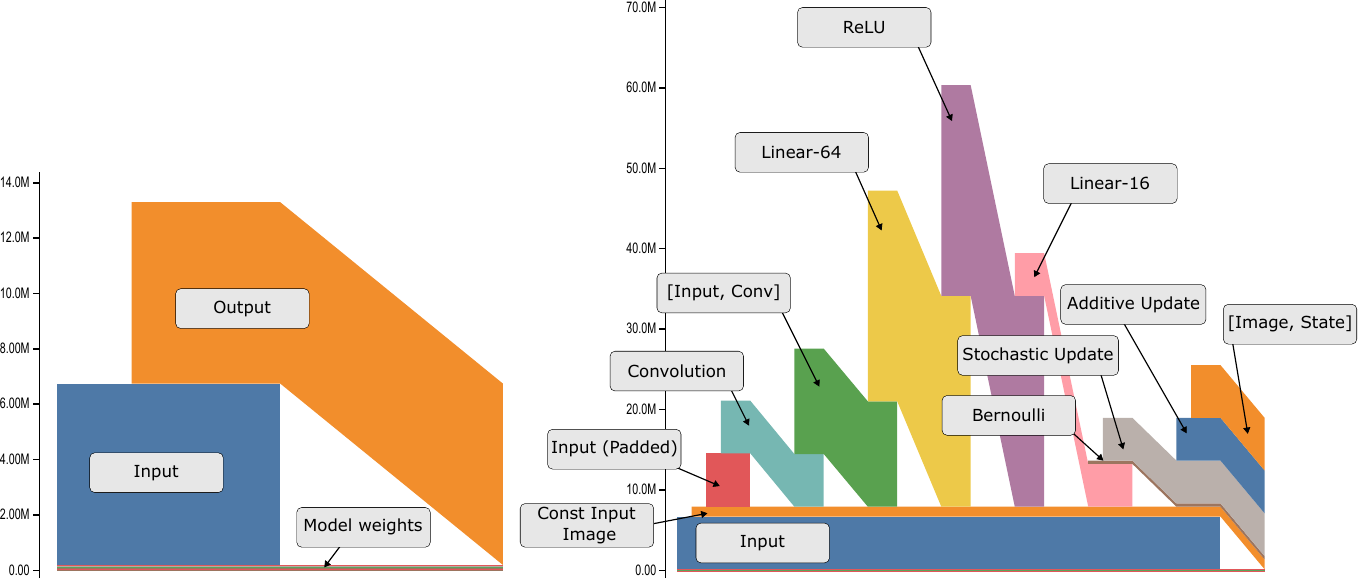}
    \caption{VRAM allocations of our CUDA kernel (left) and the traditional implementation (right).}
    \label{fig:real_mem_allocations}
\end{figure*}

\Cref{fig:real_mem_allocations} shows real memory allocations obtained by running a single NCA step using our CUDA implementation (left) and the traditional implementation (right). The visualizations are obtained using Pytorch's \texttt{torch.cuda.memory} API. 

Our CUDA implementation is comparably simple to understand. There is the input tensor (blue) and an output tensor (orange). The model weights can be seen at the very bottom of the plot (different colors for different layers). After computing the inference step the input tensor is removed from the memory.

On the right side, we show the memory usage of the traditional NCA implementation. At first glance, the real memory utilization is much more complex than what we showed in \cref{fig:cuda_implementation} in the main paper and more complicated than that of our CUDA implementation. For example, the ReLU is not computed in place, which further increases the maximum VRAM allocation. In detail, the memory allocations are as follows:

\textbf{Input:} The input seed, the NCA operates on.
\textbf{Const Input Image:} The input image, that should be left constant during the forward pass.
\textbf{Input (Padded):} Apparently, Pytorch first creates a padded representation of the input image, which is then convolved by the convolution operation.
\textbf{Convolution:} The output of the convolution operation.
\textbf{[Input, Conv]:} The initial input and the output of the convolution are concatenated to be fed into the linear layer.
\textbf{Linear-64:} The output of the first linear layer is computed.
\textbf{ReLU:} The ReLU activation is computed.
\textbf{Linear-16:} The output of the second linear layer is computed.
\textbf{Bernoulli:} A random tensor is sampled from a Bernoulli distribution to compute the additive update.
\textbf{Stochastic Update:} The multiplication of the Bernoulli tensor and the output of the linear layer. The result contains zero in the positions determined by the Bernoulli distribution.
\textbf{Additive Update:} Compute the additive update by adding the input and the result of the stochastic update.
\textbf{[Image, State]:} Concatenate the constant image and additive update (the new state). This will be the input for the next NCA step.

\section{Advantages of our NCA Inference CUDA Kernel}
As mentioned, our CUDA kernel for NCA inference has a much lower VRAM demand by storing intermediate results in thread-local registers instead of the GPU's shared main memory.
However, our implementation offers further advantages not highlighted so far:

1) The \textbf{overhead of switching between Python and C++ Code} is reduced. Especially, as Python code is executed much slower than compiled C++ Code, this can make a huge difference in terms of energy efficiency and run time.

2) The \textbf{overhead of launching CUDA kernels} is only applied once per step. Our kernel performs the entire computation in a single kernel, while traditional computations need to launch at least 5 kernels to perform a single NCA step.

3) \textbf{Less input and output operations} on the slow shared memory. Reading and writing to the shared memory is much slower compared to accessing the thread-local registers. Our kernels keep variables close to the thread operating on the cells, which reduces input and output operations on the shared memory.

4) \textbf{Understanding and debugging memory allocations}. A proper understanding of internal memory allocations of the deep learning framework is important to optimize the memory usage of inference scripts. Pytorch provides a great API for visualizing and debugging memory allocations. However, memory allocations of a traditional NCA implementation are difficult to understand and debug. Since our optimized CUDA kernel removes all unnecessary memory allocations, the memory usage is much easier to grasp and optimize. \cref{fig:real_mem_allocations} shows the memory allocations of our implementation and the allocations using the traditional implementation. The memory allocations are much more difficult to understand than those of our implementation.

\section{Pytorch Details for Reducing VRAM Demands}

\begin{figure*}
    \centering
    \includegraphics[width=\linewidth]{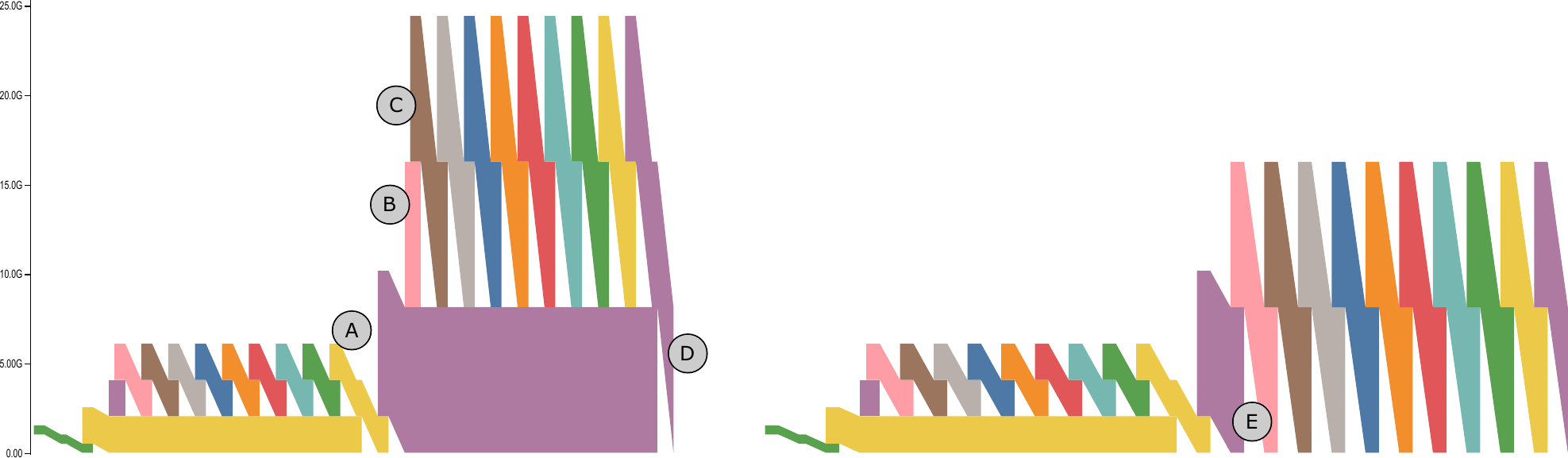}
    \caption{NCA inference invoked in a separate function (left) and in the local context (right).}
    \label{fig:forward_in_function}
\end{figure*}
Python manages the allocated memory on its own. In our Python implementation, memory is released as soon as there is no further reference to it. Since the code is interpreted, the compiler does not look ahead to check if a variable is going to be used again. The same happens with Pytorch tensors. This creates a massive problem in our implementations: Passing an input seed into a function for inference will not delete the initial seed until the function invocation is completed. The seed, which usually has a large size (compared to a simple image), unnecessarily resides in the VRAM.

In our OctreeNCA implementation, the seed is the state variable computed on the previous octree level. A simplified implementation is given in \cref{lst:inference_w_func}.

\begin{lstlisting}[caption=Invoking NCA inference in a function., label={lst:inference_w_func}, float=tp]
# upscale hidden states from previous level
state = F.interpolate(state, size=octree_lvl[0], mode='nearest')
# generate input image of octree level 0
temp = F.interpolate(slide, size=octree_lvl[0])
# combine hidden states and input image 
state[0,:model.input_channels] = temp[0]
# invoke forward function
state = model.backbone_ncas[0](state)
# at this point the old state tensor is freed
\end{lstlisting}

The problem with this implementation is that the \texttt{state} variable will live until the end of the forward pass invocation in line 4 even though the variable is immediately overwritten at the end of the function invocation. 

The improved version of this code snippet performs the inference directly within the local context as shown in \cref{lst:inference_wo_func}.

\begin{lstlisting}[caption=Invoking NCA inference in the local context., label={lst:inference_wo_func}, float=tp]
# same as before: create next octree level
state = F.interpolate(state, size=octree_lvl[0], mode='nearest')
temp = F.interpolate(slide, size=octree_lvl[0])
state[0,:model.input_channels] = temp[0]

bb = model.backbone_ncas[0] #backbone

for step in range(bb.num_steps):
    # pre-allocate output
    new_state = torch.zeros_like(state)
    # use Pytorchs random number generator
    new_state[:, 0].bernoulli_(0.5)
    # perform NCA step, write results to new_state
    nca2d_cuda(new_state, state, bb.weights)
    state = new_state #override -> free memory
\end{lstlisting}

The problem can be understood when visualizing the VRAM allocations. We use Pytorch's memory visualization methods and present the results of both implementations in \cref{fig:forward_in_function}.
In the first version (left), the final NCA is applied in a function. In the right version, the final NCA is applied within the current context, to trigger an early deletion of the \texttt{state} variable. In detail, those are the important key points to notice in this Figure:
A) The state tensor from the previous octree level is upsampled.
B) An output tensor is created and its values are computed using our optimized CUDA kernel. The result of (A) serves as the input.
C) The result of the second NCA step is computed. The state tensor from the outer context resides in the GPU VRAM.
D) The final state tensor is returned by the NCA's forward function. Overriding the \texttt{state} variable removes the final reference to the internal data, finally, triggering the deletion of the data.
E) Since the inference is invoked in the current context, overwriting the \texttt{state} variable will immediately free the memory.

This fact should be considered when optimizing the memory consumption in a Python script. Ideally, an implementation similar to \cref{lst:inference_wo_func} should be used when optimizing the VRAM demands.

\section{Training Details}
\begin{table*}[t]
    \centering
\adjustbox{max width=\linewidth}{%
    \begin{tabular}{r|lllll}
         \toprule
         Task & \multicolumn{5}{c}{Resolution at octree level}\\
         & 5&4&3&2&1\\
         \midrule
         Radiology& $320\times320\times24$ & $160\times160\times12$& $80\times80\times6$& $40\times40\times6$& $20\times20\times6$\\
         Pathology& $320\times320$ & $160\times160$ & $80\times80$& $40\times40$& $20\times20$\\
         Surgery &  $240\times432\times80$ & $120\times216\times40$& $60\times108\times20$& $30\times54\times10$& $15\times27\times5$\\
         \bottomrule 
    \end{tabular}}
    \caption{The resolution on each level of the octree structure.}
    \label{tab:octree_lvls}
\end{table*}
% EMA, loss function, iterations per epoch, patch sizes, patches in pathology task, learning rate, LR scheduler, optimizer.
In this section, we mention all the hyperparameters crucial for reproducing our work.

\textbf{Backbone NCA:} Our backbone NCA does not use any normalization. We use a fire rate of $50\%$. The kernel size of each NCA is $3\times3(\times3)$. The NCA has $16$ state channels and a hidden size of $64$. We use the same architecture, but different weights for each resolution. 

\textbf{Octree:} For each task, we construct an octree data structure of five levels. The resolution of each level is found in \cref{tab:octree_lvls}. During training on the radiology and surgical data, we extract patches on the two lowest octree levels to reduce the VRAM demands. On the radiology data, we train on $80\times80\times6$ patches. On the surgery data, we use $60\times106\times20$ patches.

\textbf{Training:} We train our models using the Adam optimizer with parameters $\beta_1=0.9$ and $\beta_2=0.99$. The base learning rate is $1.6\cdot10^{-3}$ and is scheduled by an exponential learning rate with a decay rate of $\delta=0.9992$. To compensate for noisy training, we employ an exponential moving average which updates the parameters of a shadow model $\theta_\text{S}$ according to
\begin{align*}
    \theta_\text{S} \gets \alpha \cdot\theta_\text{S} + (1-\alpha) \cdot\theta
\end{align*}
where $\alpha\in[0, 1)$ is the decay parameter and $\theta$ are the parameters of the currently updated model. We choose $\alpha = 0.99$ for all our models. Evaluation is performed with the shadow model that does not suffer from noisy gradient updates. We train the models using a combination of a cross-entropy loss and Dice loss with weight $\lambda_\text{Dice}$:
\begin{align*}
\loss=(2-\lambda_\text{Dice})\loss_\text{BCE}+\lambda_\text{Dice}\loss_\text{Dice}
\end{align*}
On the radiology and pathology data, we train using a batch size of $3$, while on the surgery dataset, we use a batch size of $1$. Since the pathology slices are too large, we train on $320\times320$ patches. We exclude patches that do not contain any foreground (epithelium cells). This results in more than $77,000$ patches, which would be cumbersome to include in each epoch. Hence, we restrict one epoch to 200 random batches. 
We implement our model and experiments in the M3D-NCA framework~\cite{kalkhof2023m3d}. We further implement the CUDA NCA inference function in CUDA and C++.

\section{Baselines}
% UNet, nnUNet, UNet++, (Densenet, Resnet18, VGG11, Movilenetv2), SegFormer, SAM (including the weights), SAM2 (including the weights) (SAM & SAM2 inference procedure: How we place labels!)
We compare our OctreeNCA to different segmentation baselines, which we introduce here in more detail.

\textbf{UNet:}
% Convolutions, pooling, up-convolutions
The UNet~\cite{ronneberger2015u} is a fully convolutional segmentation model consisting of an encoder and a decoder. The encoder consists of a series of convolutions and pooling operations. The decoder consists of convolutions and upsampling operations. The UNet has an equal amount of downsampling (pooling) and upsampling operations, resulting in a symmetric architecture. Skip connections from the encoder provide high-resolution details to the decoder.

\textbf{nnUNet:} The nnUNet~\cite{isensee2019nnu} framework is an auto ML pipeline that automates the pre-processing and planning of the model and training hyperparameters. The underlying architecture is a UNet.

% densly connected nested decoder sub-networks
% "bridge the semantic gap between the feature maps of the encoder and decoder prior to fusion"
\textbf{UNet++:} The UNet++~\cite{zhou2018unet++} extends the UNet with nested decoder networks. The intermediate decoders bridge the semantic gap between the activations in the encoder and decoder.

\textbf{VGG11:} The VGG11~\cite{simonyan2014very} contains 11 weight layer. Removing the linear layers leaves the VGG11 with 8 convolutional layers and 5 pooling layers, which create feature activations at different resolutions. Combined with a UNet decoder it makes a fully convolutional network capable of image segmentation.

\textbf{Resnet18:} The Resnet18~\cite{he2016deep} introduces residual connections for dealing with vanishing gradients in image classification. Combined with a UNet decoder, it can be used for segmentation as well.

\textbf{Densenet:}  The Densenet~\cite{huang2017densely} proposes dense blocks that, similarly to Resnets, take in features from previous layers. However, they are not directly added, as in ResNet18, but concatenated.

\textbf{Mobilenetv2:} % Residual Connections + sophisticated combination of depthwise and pointwise convolutions
The Mobilenetv2~\cite{sandler2018mobilenetv2} is a sophisticated architecture containing residual connections and a combination of depthwise and pointwise convolutions.

\textbf{Efficientnet:} Based on the Mobilenetv2, the Efficientnet~\cite{tan2019efficientnet} provides an architecture that is easily scalable using a single parameter jointly controlling the width and the depth of the model.

\textbf{SegFormer:}
% hierarchical transformer encoder
% MLP decoder to predict a dense segmentation
The SegFormer~\cite{xie2021segformer} is a vision transformer tailored towards semantic segmentation. The encoder consists of a hierarchical transformer encoder. The decoder consists of fully connected layers to predict a dense segmentation mask.

\textbf{SAM:} The Segment Anything Model (SAM)~\cite{kirillov2023segment} is a foundational model consisting of an image encoder, a prompt encoder, and a mask decoder. The model is trained for promptable segmentation, meaning that together with the image a prompt must be given to the model to deal with ambiguity present in the image. SAM is available in different sizes. In our experiments, we use the pre-trained ViT-H backbone. We prompt the model by 20 random foreground points in each $320\times320$ pathology patch. SAM can only segment 2D images, so we do not apply it to the 3D radiology MRIs or the surgical videos.

\textbf{SAM2:} SAM2~\cite{ravi2024sam2} is an updated version of SAM, that allows the segmentation of videos. SAM2 augments SAM with a memory bank and an attention memory. The memory bank stores information from previous frames, which is reused for segmenting subsequent frames. We use the pre-trained weights of the Hiera-Base+ variant. We prompt the model with a single sample per class in the frame of its first occurrence.

\section{Large Scale Inference Results}\label{sec:results:large_scale}
\begin{table}[t]
    \centering
%\input{tables/single_inference_table}
%\begin{tabular}{l|rr|rr}
%\toprule
%& \multicolumn{2}{c|}{Size} & \multicolumn{2}{c}{Size / Time}\\
%& Path. & Surgery & Path. & Surgery\\
%\midrule
%UNet            &8.3& 3.8 &9.76&4.47\\
%med/M3D-NCA     &28.7& 11.5 &7.79&4.32\\
%OctreeNCA       &28.7& 11.5 &13.93&8.77\\
%OctreeNCA-CUDA     &184.9& 76.2 & 77.36 &18.14\\
%\bottomrule
%\end{tabular}

\begin{tabular}{l|rr|rr}
\toprule
& \multicolumn{2}{c|}{Pathology} & \multicolumn{2}{c}{Surgery}\\
& MP& MP/s & sec. & FPS\\
\midrule
UNet            &8.3&9.76& 3.8 &112\\
med/M3D-NCA     &28.7&7.79& 11.5 &108\\
OctreeNCA       &28.7&13.93& 11.5 &219\\
OctreeNCA-CUDA     &\textbf{184.9}& \textbf{77.36}& \textbf{76.2} &\textbf{453}\\
\bottomrule
\end{tabular}
    \caption{The maximum inference size applicable using an Nvidia RTX 4090 24GiB GPU and the corresponding runtime. Inference size is given in Megapixel for pathology segmentation and in seconds (sec.) for surgery segmentation.}
    \label{tab:max_inference_gpu}
\end{table}
In \cref{tab:max_inference_gpu}, we investigate the maximum size of a pathology slice and the maximum length of a cholec video processable in a single pass on an Nvidia RTX 4090 GPU with 24 GiB VRAM. The UNet can only segment a measly 8.3 Megapixel (MP) image or a 3.8-second-long video. Med-NCA, respectively M3D-NCA, segments 28 MP patches or 11.5-second videos at once. Our OctreeNCA has a similar VRAM demand so, the maximum applicable size is the same for our method. However, our OctreeNCA inference is almost twice as quick. If we utilize our custom CUDA implementations (indicated by OctreeNCA-CUDA), we can segment more than 6 times larger inputs. A pathology slice with 184 MP or a more than one-minute-long surgical video is no longer a problem for our OctreeNCA. The inference is carried out tremendously quickly, being at least 4 times quicker than the quickest baseline.

\section{Ablation Study and Comparison to M3D-NCA}\label{sec:results:ablation}
\begin{table}[]
    \centering
    \begin{tabular}{cc|cc|cc|cc}
    \toprule
       \multicolumn{2}{c|}{Task /} & \multicolumn{2}{c|}{Radio.} & \multicolumn{2}{c|}{Path.} & \multicolumn{2}{c}{Surgery}\\
       \multicolumn{2}{c|}{Ablation}&D&M&D&M&D&M\\
    \midrule
    \parbox[t]{2mm}{\multirow{3}{*}{\rotatebox[origin=c]{90}{Norm}}}
         &\textbf{w/o}	&  \best79.2 & \best2.2 & \sbest86.3 & \best2.2 & 56.4 & \best7.2 \\
         &Batch			& \sbest78.1 & 3.2 & 84.5 & 3.3 & \best63.7 & 10.1 \\
         &Layer			&      66.7 & 3.2 & \best86.5 & 3.3 & \sbest56.8 & 10.1 \\
    \midrule
    \parbox[t]{2mm}{\multirow{3}{*}{\rotatebox[origin=c]{90}{Steps}}}
         &5				& \best80.0 & \best1.3 & 85.6 & \best1.3 & 55.3 & \best4.2 \\
         &\textbf{10}	&  79.2 & \sbest2.2 & \sbest86.3 & \sbest2.2 & \sbest56.4 & \sbest7.2 \\
         &20			    &\sbest79.8 & 4.2 & \best86.6 & 4.2 & \best58.5 & 13.1 \\
    \midrule
    \parbox[t]{2mm}{\multirow{3}{*}{\rotatebox[origin=c]{90}{Step $\alpha_0$}}}
         &\textbf{1.0}	& 79.2 & \best2.2 & \sbest86.3 & 2.2 & \best56.4 & \best7.2 \\
         &1.5			& \best82.3 & \sbest2.3 & 86.2 & \best2.2 & \sbest55.1 & \sbest7.2 \\
         &2.0			&\sbest80.3 & 2.3 & \best86.3 & \sbest2.2 & 54.4 & 7.3 \\
    \midrule
    \parbox[t]{2mm}{\multirow{3}{*}{\rotatebox[origin=c]{90}{Model}}}
         &12/48					&   76.3 & \best1.7 & 85.8 & \best1.7 & 53.0 & \best5.6 \\
         &\textbf{16/64}		&  \best79.2 & \sbest2.2 & \sbest86.3 & \sbest2.2 & \sbest56.4 & \sbest7.2 \\
         &20/100				& \sbest78.4 & 3.0 & \best86.6 & 3.1 & \best64.1 & 9.6 \\
    \midrule
    \parbox[t]{2mm}{\multirow{3}{*}{\rotatebox[origin=c]{90}{Batch}}}
         &1					& \best81.8 & \best0.8 & 85.6 & \best0.9 & \best65.0 & \best2.4 \\
         &\textbf{3}		&  79.2 & \sbest2.2 & \sbest86.3 & \sbest2.2 & \sbest56.4 & \sbest7.2 \\
         &5					&\sbest79.5 & 3.6 & \best86.4 & 3.6 & 56.1 & 11.9 \\
    \midrule
    \parbox[t]{2mm}{\multirow{4}{*}{\rotatebox[origin=c]{90}{$\lambda_\text{Dice}$}}}
         &0.0			&       75.0 & 2.2 & 86.1 & 2.2 & \best57.2 & 7.2 \\
         &\textbf{1.0}	& \sbest79.2 & 2.2 & 86.3 & 2.2 & 56.4 & 7.2 \\
         &1.2			&  \best80.7 & 2.2 & \best86.3 & 2.2 & 56.9 & 7.2 \\
         &1.5			&  77.9 & 2.2 & \sbest86.3 & 2.2 & \sbest57.0 & 7.2 \\
    \midrule
    \parbox[t]{2mm}{\multirow{3}{*}{\rotatebox[origin=c]{90}{EMA}}}
         &0.0			&       74.9 & 2.2 & 86.1 & 2.2 & 56.2 & 7.2 \\
         &0.9			& \sbest78.8 & 2.2 & \best86.3 & 2.2 & \best60.4 & 7.2 \\
         &\textbf{0.99}	&  \best79.2 & 2.2 & \sbest86.3 & 2.2 & \sbest56.4 & 7.2 \\
    \bottomrule
    \end{tabular}

    \caption{\textbf{D}ice (\%) and VRAM \textbf{M}emory utilization in GiB of the GPU during training of OctreeNCA using different normalization layers, different number of steps per upsampling layer, model size in num. channels / hidden layers, training batch size, the Dice loss weight, and the Exponential Moving Average (EMA) decay. Results are for Radiology, Pathology, and Surgery segmentation.}
    \label{tab:ablation_norm}
\end{table}
\begin{table}[]
    \centering
\adjustbox{max width=\linewidth}{\begin{tabular}{c|rrr|rrr}
    \toprule
         \textbf{L} & \multicolumn{3}{c|}{M3D-NCA} &\multicolumn{3}{c}{OctreeNCA}\\ 
         & Dice & \textbf{Mem} & \textbf{P} & Dice & \textbf{Mem} & \textbf{P}\\
    \midrule
         2 & 80.5 & 14.34 & 12,480 & 77.3 & 10.04 & 7,040\\
         3 & 81.8 & 14.49 & 16,000 & 77.6 & 5.98 & 10,560\\
         %4 & 79.2 & 11.08 & 16,160 & 77.3 & 2.50 & 14,080\\
         5 & 80.9 & 11.11 & 19,808 & 82.5 & 2.18 & 17,600\\
    \bottomrule
\end{tabular}}
    \caption{Dice, \textbf{Mem}ory consumption during training (in GiB) and the number of \textbf{P}arameters of M3D-NCA and OctreeNCA with different number of \textbf{L}evels on the radiology segmentation task.} 
    \label{tab:m3d_vs_oct}
\end{table}
We perform a thorough analysis of the design choices of our OctreeNCA. The results are presented in \cref{tab:ablation_norm}. We perform ablation over different normalization layers, the number of steps for each octree level $l>0$. We test different numbers of steps on the first octree level depending on the resolution of the octree level computed as $\alpha_0\cdot\max(H,W,D)$ where $H,W$ and $D$ are the dimensions of the input. We test different NCA capacities jointly with different numbers of hidden channels. The table references models in $c/h$ where $c$ is the number of channels and $h$ is the hidden size of the NCA.
We evaluate design choices on batch size, the weight of the dice loss $\lambda_\text{Dice}$ and the Exponential Moving Average (EMA) decay.
The standard model configuration is indicated by the bold design choices.

In \cref{tab:m3d_vs_oct} we ablate on the number of Octree levels on the radiology segmentation task only. Additionally, we compare OctreeNCA to M3D-NCA using the same number of levels. 
Increasing the number of levels has a non-significant influence on the Dice score and memory allocation for M3D-NCA. Contrary, OctreeNCA benefits in terms of Dice score and memory allocation when increasing the number of layers, while requiring fewer parameters.

\begin{table}[t]
    \centering
    \begin{tabular}{lrrr}
\toprule
Model & VRAM& Num. Params & Dice \\
\midrule
OctreeNCA & 2.18 & 17,600 & 82.51 \\
M3D-NCA & 14.34 & 12,480 & 80.52 \\
UNet & 7.79 & 19,071,297 & 84.06 \\
Resnet18 & 4.30 & 42,611,121 & 85.62 \\
Densenet & 6.20 & 31,199,985 & 84.76 \\
nnUNet & 7.35 & 44,797,408 & 88.18 \\
\bottomrule
\end{tabular}

    \caption{VRAM demand during training (in GiB), the number of parameters, and the Dice score on the radiology segmentation task.}
    \label{tab:supp_prostate}
\end{table}
\begin{table}[t]
    \centering
    \begin{tabular}{lrrr}
\toprule
Model & VRAM& Num. Params & Dice \\
\midrule
OctreeNCA & 2.25 & 15,520 & 86.31 \\
Med-NCA & 6.87 & 7,488 & 84.41 \\
UNet & 5.49 & 36,951,425 & 82.79 \\
Efficientnet & 2.68 & 6,251,469 & 88.15 \\
Mobilenetv2 & 2.47 & 6,628,945 & 88.20 \\
Resnet18 & 1.56 & 14,328,209 & 88.00 \\
VGG11 & 2.50 & 18,254,033 & 87.66 \\
UNet++ & 3.86 & 26,078,609 & 88.23 \\
SegFormer & 0.50 & 3,714,658 & 86.31 \\
SAM & - & 641,090,608 & 41.12 \\
\bottomrule
\end{tabular}

    \caption{VRAM demand during training (in GiB), the number of parameters, and the Dice score on the pathology segmentation task.}
    \label{tab:supp_peso}
\end{table}
\begin{table}[t]
    \centering
\adjustbox{max width=\linewidth}{%
    \begin{tabular}{lrr|rrrrr|r}
\toprule
&VRAM&Num.&Abdominal&&&&&\\
Model & (train) & Params. & wall & Liver & Fat & Grasper & Gallbladder & mDice \\
\midrule
OctreeNCA & 2.41 & 16,960 & 70.62 & 76.92 & 74.64 & 46.23 & 56.68 & 64.01 \\
M3D-NCA & 13.33 & 12,480 & 71.85 & 74.12 & 77.78 & 45.64 & 54.41 & 64.03 \\
UNet & 13.96 & 19,073,285 & 59.22 & 58.97 & 66.47 & 45.07 & 41.82 & 53.13 \\
Resnet18 & 16.08 & 42,656,757 & 74.89 & 72.82 & 70.46 & 40.93 & 54.51 & 61.90 \\
Densenet & 22.44 & 31,245,621 & 70.51 & 74.25 & 76.82 & 42.62 & 55.42 & 63.33 \\
SAM2 & - & 80,833,666 & 57.04 & 53.44 & 49.49 & 82.29 & 66.20 & 60.07 \\
\bottomrule
\end{tabular}
}
    \caption{VRAM demand during training (in GiB), the number of parameters, per-class Dice scores, and the mean Dice (mDice) on the surgical video segmentation task.}
    \label{tab:supp_cholec}
\end{table}

\section{Detailed Quantitative Results}\label{sec:results:detailed_quantitative}

\textbf{Radiology: } \Cref{tab:supp_prostate} shows the results obtained on the radiology segmentation task. We observe that our OctreeNCA requires the least amount of VRAM during training. Especially compared to M3D-NCA, the other NCA-based method, our OctreeNCA reduces VRAM demands by $84.7\%$. The NCA-based methods have fewer parameters than the CNNs. Our OctreeNCA outperforms M3D-NCA in terms of segmentation accuracy.

\textbf{Pathology:} We report the detailed pathology results in \cref{tab:supp_peso}. In terms of GPU allocations during training, most methods have moderate VRAM demand. Only the Med-NCA and the UNet have a comparably high VRAM demand. The SegFormer has a rather low VRAM demand. However, due to the quadratic scaling, the VRAM demand during inference, on larger images would be much larger (\cref{fig:raspi_speed_and_gpu_vram}). As expected, the NCA-based methods have the smallest number of parameters. The foundation model SAM has by far the most parameters - almost $20\times$ as many parameters as the UNet. However, the Dice score of SAM is very low, indicating that SAM does not work well on pathology images. The other methods perform reasonably well, with only minor differences between each other.

% OctreeNCA least VRAM, slightly more parameters than M3D-NCA, but way less parameters than CNNs or SAM2

\textbf{Surgery:} The results are found in \cref{tab:supp_cholec}. Since the surgery dataset features multiple classes, we report segmentation accuracies per class next to the mean Dice (mDice).
OcteeNCA needs less than $81\%$ of any of the baselines. It has slightly more parameters than M3D-NCA, but $1000$-times fewer parameters than the CNNs.  
OctreeNCA performs well at segmenting the abdominal wall and liver, where OctreeNCA performs best. Segmenting the Grasper and the Gallbladder is a rather difficult task, because the classes are still rare in the train set, making every trained method suffer from a rather poor performance. SAM2 performs rather well due to the prompts we give to the model. The grasper is a small object. The prompt point on the grasper already provides an excellent indication of where to find it. While our methods struggle with detecting the grasper in the first place, the prompt provides this information. Also since the grasper is an object that looks like a metal bar, the pre-trained SAM2 has probably seen many objects similar to its appearance. This advantageous performance on the grasper and the gallbladder makes SAM2 perform comparably to the other methods in terms of mean Dice.

\textbf{Timings on Raspberry Pi:} We report the exact values of our runtime analyses on the Raspberry Pi in \cref{tab:supp_pi_speed}. 

\begin{table}[]
    \centering
\adjustbox{max width=\linewidth}{%
    \begin{tabular}{l|rrrrrrrrr}
\toprule
Side length & 320 & 400 & 480 & 560 & 640 & 720 & 800 & 880 & 960 \\
 \midrule
OctreeNCA & 2.20 & 3.54 & 4.89 & 6.54 & 8.81 & 10.97 & 14.16 & 18.30 & 21.53 \\
Med-NCA & 4.41 & 7.15 & 9.99 & 13.62 & 17.87 & 22.60 & 28.25 & 35.59 & 44.20 \\
UNet & 10.09 & 16.20 & 23.19 & 32.43 & 42.15 & - & - & - & - \\
\bottomrule
\end{tabular}}
    \caption{Inference speed (in sec.) on a Raspberry Pi when applied to a quadratic image.}
    \label{tab:supp_pi_speed}
\end{table}

\section{Further Qualitative Results}\label{sec:results:more_qualitative}

\begin{figure*}[t]
    \centering
    \includegraphics[width=\linewidth]{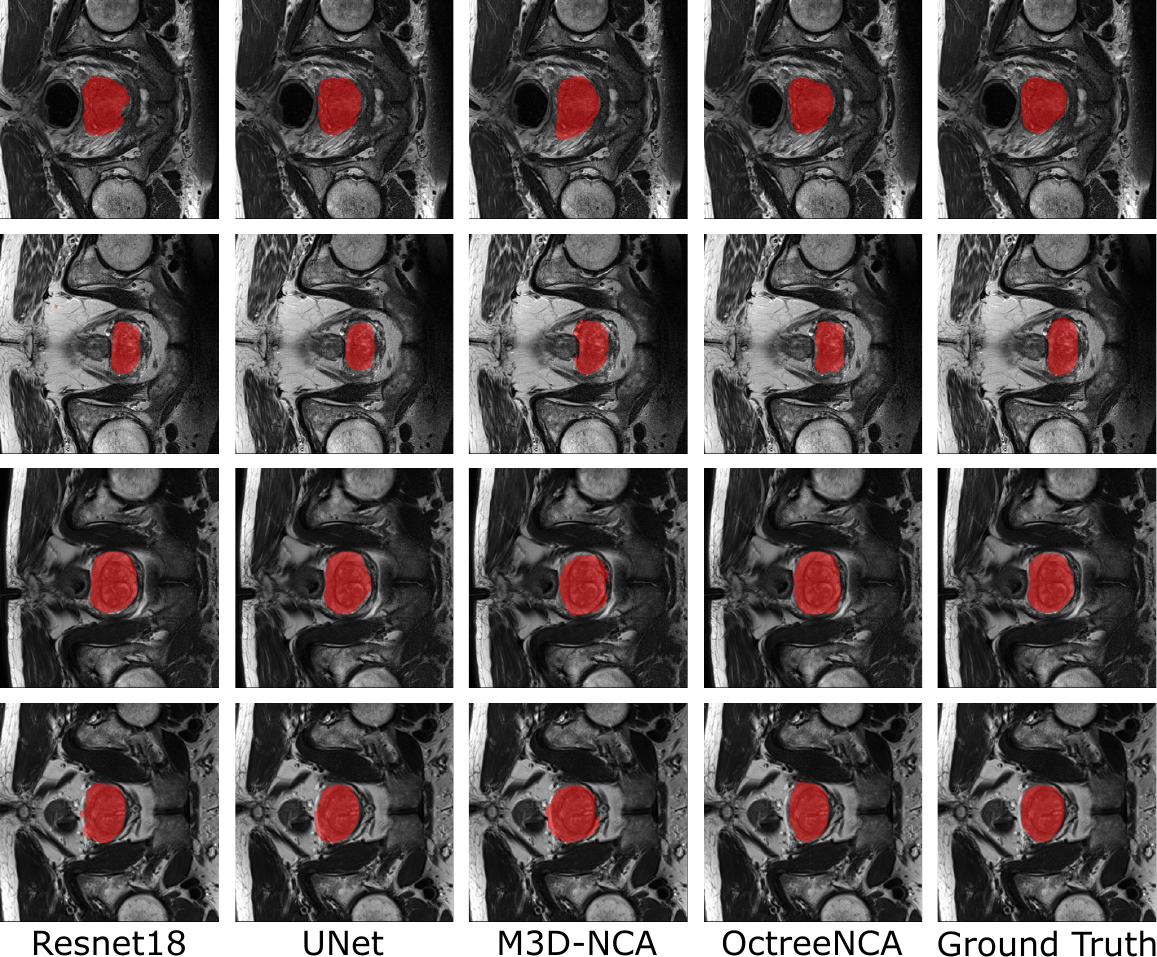}
    \caption{Middle slices from radiology segmentation of the prostate of four different patients. Segmented using Resnet18, UNet, M3D-NCA and OctreeNCA.}
    \label{fig:supp_prostate}
\end{figure*}

\begin{figure*}[t]
    \centering
    \includegraphics[width=\linewidth]{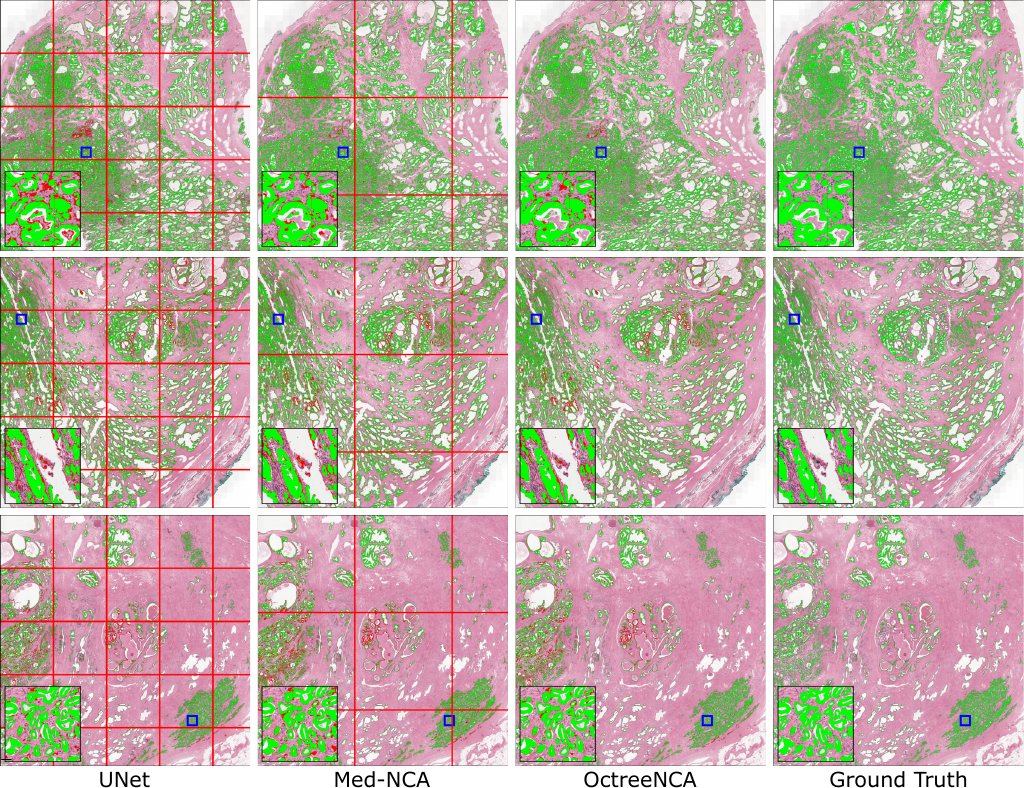}
    \caption{Pathology segmentation results of UNet, Med-NCA, and OctreeNCA on slices from three patients. True positive pixels are green, false positives are red. The UNet and Med-NCA segmentations are stitched together from the red patches. The inset shows a zoomed-in version of the blue patch.}
    \label{fig:supp_peso}
\end{figure*}

\begin{figure*}[t]
    \centering
    \includegraphics[width=\linewidth]{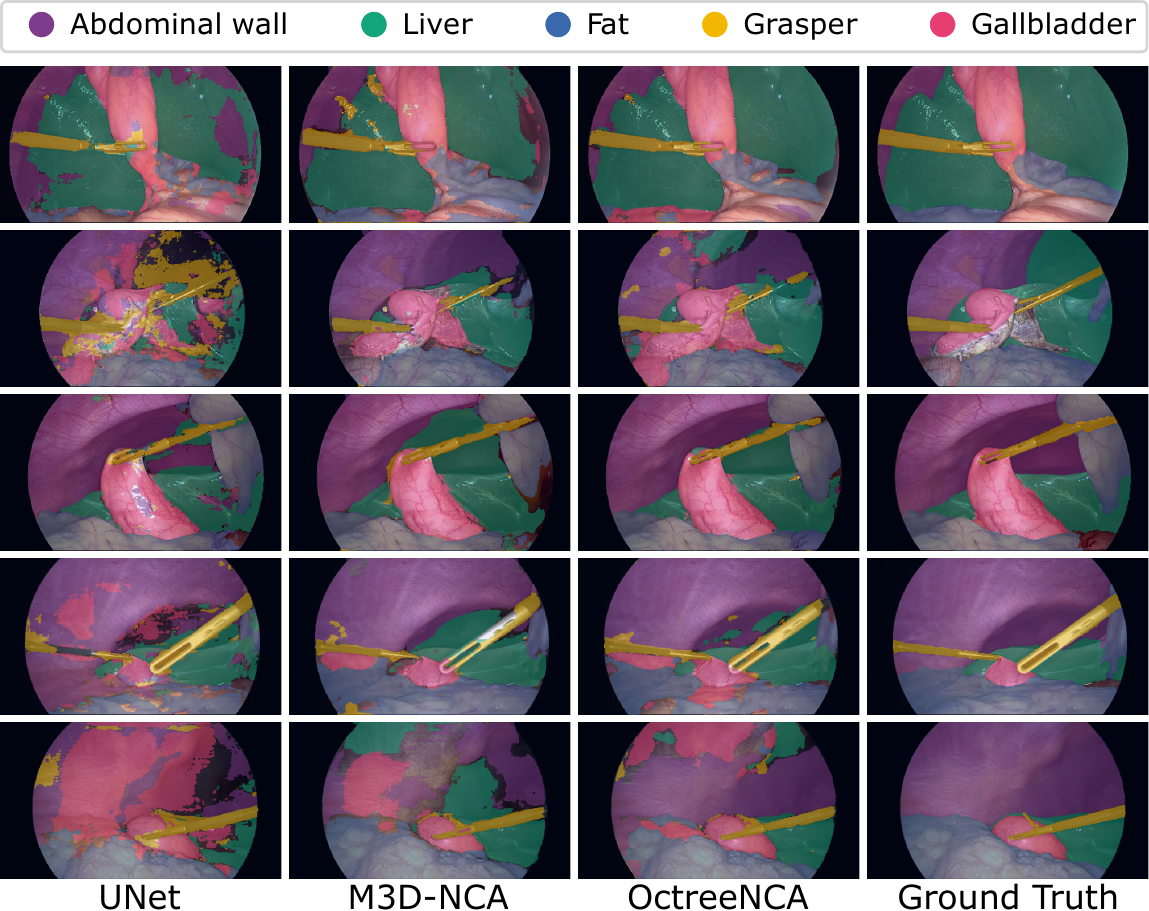}
    \caption{Frames of surgical video segmentations of UNet, M3D-NCA, and OctreeNCA of five different operations.}
    \label{fig:supp_cholec}
\end{figure*}

In \cref{fig:supp_prostate}, we show four different radiology segmentation results using different architectures.
\Cref{fig:supp_peso} shows three segmentations of pathology slices extracted from different patients.
Lastly, \cref{fig:supp_cholec} shows frames extracted from segmentations of five different operations.

\end{document}